# On Cropped versus Uncropped Training Sets in Tabular Structure Detection

Yakup Akkaya [a], Murat Simsek [a], Burak Kantarci [a], Shahzad Khan [b]

[a] School of Electrical Engineering and Computer Science, University of Ottawa, Ottawa, ON, K1N 6N5, Canada
[b] Lytica Inc., 308 Legget Dr, Kanata, ON K2K 1Y6, Canada

Abstract

Automated document processing for tabular information extraction is highly desired in many organizations, from industry to government. Prior works have addressed this problem under table detection and table structure detection tasks. Proposed solutions leveraging deep learning approaches have been giving promising results in these tasks. However, the impact of dataset structures on table structure detection has not been investigated. In this study, we provide a comparison of table structure detection performance with cropped and uncropped datasets. The cropped set consists of only table images that are cropped from documents assuming tables are detected perfectly. The uncropped set consists of regular document images. Experiments show that deep learning models can improve the detection performance by up to 9% in average precision and average recall on the cropped versions. Furthermore, the impact of cropped images is negligible under the Intersection over Union (IoU) values of 50%-70% when compared to the uncropped versions. However, beyond 70% IoU thresholds, cropped datasets provide significantly higher detection performance.

Keywords: Deep Learning, Convolutional Neural Networks, Image Processing, Document Processing, Table Detection, Tabular Data Extraction, Page Object Detection, Structure Detection.

*This work was supported in part by Mathematics of Information Technology and Complex Systems (MITACS) Accelerate Program under the Award Number IT14836, and Smart Computing for Innovation (SOSCIP) Program under Project Number 3-040, and Lytica Inc.

Email addresses: yakka087@uottawa.ca (Yakup Akkaya [a]), murat.simsek@uottawa.ca (Murat Simsek [a]), burak.kantarci@uottawa.ca (Burak Kantarci [a]), shahzad_khan@lytica.com (Shahzad Khan [b])

October 6, 2021

1. Introduction

The amount of unstructured data pushed by billions of connected devices has become overwhelmingly high in a variety of categories and forms with the advent of Industry 4.0 paradigm. Alongside, artificial intelligence, machine-to-machine communication, and Internet of Things (IoT) devices are being incorporated into automated manufacturing to minimize human intervention in production [1]. Automation of data processing and analysis is also a critical component of this cycle for many organizations that contribute to the supply chain [2].

Tables in documents with huge volumes of data often contain valuable information such as cost, technical specifications, requirements, and performance capabilities. Automated extraction of this information from documents is of paramount importance particularly to pave the way for optimized supply chain management. The state of the art offers image recognition-based approaches with Convolutional Neural Network (CNN) classifiers as a solution to this problem. However, heterogeneity of table structures and layouts remain a challenge for detection.

Extracting tabular information out of the image documents involve two separate tasks. The first step is to detect the table with the position on the document. Once the table region is captured, the structure of the table is detected. This task can be executed by segmenting the table layout as rows and columns or simply as table cells. Prior to the deep learning-based solutions, existing methods were largely operating on on digital-born PDF documents (which retained all PDF meta-data). PDF-based approaches exploit the meta-data associated with documents in combination with heuristic rules [3, 4]. However, PDF documents may not contain metadata with standardized structure since most of the documents are created from scanned images.

Deep learning based approaches have been popular for table detection and structure detection. Works that have been published on structure detection are relatively much rarer than those on table detection [5, 6, 7, 8, 9, 10, 11], due to the more complicated nature of the problem. The lack of standards for textual regions and other figures within the documents make determining table objects challenging. Recognizing the anatomy of the table is a more challenging task than table detection as it requires detecting irregularities such as spanning rows and columns, and the identification of merged cell objects in complex table structures. Rows and cells can be quite densely laid out in a page, and have tiny dimensions. Besides, both tasks require high volumes of hand labelled data, and hand-labelling is a time-consuming task, and thus, these useful training resources are scarce.

Many studies have adopted Region-Based Convolutional Neural Network (R-CNN) algorithms

in table detection and structure detection. The presented works in [12, 13] achieved satisfactory results with over 96% F1-score for row and column detection by using Mask R-CNN. Prasad et al. [14] utilized Cascade Mask R-CNN, which is developed upon Mask R-CNN and achieved the highest accuracy results on the ICDAR 2019 structure detection dataset. Jiang et al. [15] show that Hybrid Task Cascade with dual ResNeXt101 backbone outperforms the existing solutions by providing around 8%-9% higher F1-Score (81.8%) than Mask R-CNN and Cascade Mask R-CNN in table structure detection task in cell recognition.

In this study, we present the performance effect of cropped and uncropped training datasets on table structure detection. The uncropped dataset consists of regular document images containing tables whereas the cropped dataset is formed with only table images cropped from whole document images. Eliminating surrounding objects such as plots, figures and text regions in the search process of table structure, prevents false-positives and improves model performances. Models trained with cropped images can achieve higher average precision (AP), average recall (AR) and average F1-score (AF1) which is an increase of 9% over the models trained with uncropped images. Remarkably, cropped sets provide better detection results at higher IoUs of between 80% and 90%, while either set can perform better or similarly at lower IoU values. This phenomenon is vitally important as industrial applications require higher granular accuracy to be adopted, forming the basis for the motivation of this research.

The rest of the article is organized as follows: Section 2 summarizes the related works on table structure detection. Section 3 explains the methodology and model details. In section 4, the performance comparison of cropped and uncropped datasets was made by using different networks. Finally, Section 5 concludes the article.

## 2. Related Work and Motivation

### 2.1. Motivation

Table detection and table structure detection tasks form the process of extracting tabular information from documents in the literature. Existing studies either present an end-to-end approach to perform both tasks or directly focus on structure detection. Most structure detection related studies assume that tables are perfectly localized and focused on the detected table area. However, the impact of having a table detection model on the performance of structure recognition task

has not been discussed in the literature. In other words, a model can seek table structure in the whole document image instead of the detected table area. Thus deep learning model might generalize better on table structure by learning with surrounding textual regions and other figures. Consequently, a robust structure detection model can make the table detection step impractical. On the other hand, regardless of the robustness of a structure detection model, the table detection step may provide additional improvement by restricting the search area to the detected region. Therefore a comparison between the two approaches is beneficial.

2.2. State of the Art

Table content is extracted in two stages: 1) table detection, 2) table structure recognition. Far more research studies focus on table detection alone, and there are very few papers that describe work done to perform table structure recognition. In order to carry out these studies, there are mainly two approaches: 1) PDF-based, 2) image-based. PDF-based solutions use metadata associated with documents by utilizing visual indicators such as ruling lines, white spaces and vertical and horizontal alignments [3, 4]. Although early proposed works apply heuristics in image-based solutions, they heavily rely on machine learning and deep learning techniques with their growing popularity and effectiveness in automated document processing. In this section, we present works that adopt image-based approaches and perform table structure recognition tasks or both table detection and structure recognition tasks by using deep learning techniques.

Schreiber et al. [16] propose a deep learning based model for Faster-RCNN-based table detection and Fully Convolutional Network (FCN)-based structure recognition to segment the rows and columns in detected table regions. Siddiqui et al. [17] propose to use a combination of deformable convolutional networks with Faster R-CNN and FPN. With ResNet-101 as a deformable base model, deformable convolutions solve the issue of fixed receptive field by using extra offsets. These offsets enable the network to change its receptive field depending on the position and object of each input. Paliwal et al. [18] proposes combining table detection and column detection tasks by using FCN architecture with VGG-16 as the backbone. After table regions and columns are determined, rows are extracted by using Tesseract OCR with heuristic rules. The authors in [12] empower Mask-RCNN to detect rows and columns. In an extended version of that work [13], a holistic system is presented with the following components: table detection, structure detection and an end-to-end table and structure detection model with an additional judging mechanism for validation of table detections. Prasad et al. [14] present CascadeTabNet that detects tables with their types as bordered and borderless by utilizing Cascade-Mask-RCNN with High Resolution Network (HRNet). The Deep Learning model is used to accomplish cell structure recognition for borderless tables, whereas the line detection algorithm is used for bordered tables with post-processing in both branches. Jiang et al. [15] presented a benchmark on ICDAR 2013 dataset for cell structure detection by using state-of-the-art object detectors in combination with different backbones. Hybrid Task Cascade (HTC) with CBNet double ResNeXt101 outperformed the compared models.

Existing solutions for automated tabular information extraction from documents address the problem under table detection and structure detection tasks. They are compared based on various criteria in Table 1.

Table (1) Comparison of existing solutions

| Method | Training Dataset | Testing Dataset | Cropped/Uncropped Dataset | Table Detection | Structure Recognition | Pre-processing or Post-processing | Detected Element | Evaluation Metric |
|---|---|---|---|---|---|---|---|---|
| Schreiber et al. [16] | ICDAR 2013 | ICDAR 2013 (Test Set) | N/A | ✓ | ✓ | ✓ | Row/Column | Adjacency [19] relations |
| Siddiqui et al. [17] | ICDAR 2013 Custom ICDAR 2017 | ICDAR 2013 (Complete) ICDAR 2013 (Test Set) | N/A | × | ✓ | × | Row/Column | Adjacency relations |
| Paliwal et al. [18] | Marmot | ICDAR 2013 (Test Set) | Uncropped Dataset | ✓ | ✓ | ✓ | Row/Column/Cell | Adjacency relations |
| Kara et al. [12] | UNLV | UNLV | Cropped Dataset | × | ✓ | ✓ | Row/Column | IoU |
| Kara et al. [13] | UNLV | UNLV Private Dataset | Cropped Dataset | ✓ | ✓ | ✓ | Row/Column | IoU |
| Jiang et al. [15] | ICDAR 2013 | ICDAR 2013 (Test Set) | Cropped Dataset | × | ✓ | × | Cell | IoU |
| Ours | ICDAR 2013 ICDAR 2017 | ICDAR 2013 (Test Set) ICDAR 2013 (Complete) | Cropped and Uncropped Dataset | × | ✓ | × | Cell | IoU |

3. Methodology

This section presents the cropped and uncropped methods to perform structure detection under two sets of experiments. It is worth to note that developing a network architecture is not the focus of this paper; however, the architectures of already implemented networks are used in table structure detection.

3.1. The Cropped and Uncropped Approaches

It is assumed that table location is perfectly determined by a table detection model to perform structure detection task in state of the art. We adopt this approach by cropping tables from document images to perform structure detection task. This process is illustrated in Fig. 2. On the other hand, document images have plots, figures or textual areas surrounding tables. These non-table objects either may help the model to generalize better on table structure or degrade the performance. These document images are referred to as uncropped or regular, and the structure detection process for this kind of image is summarized in Fig. 2. We conduct two sets of experiments to compare the impact of cropped and uncropped sets on table structure detection performance. First, only ICDAR 2017 dataset is used for the comparison of cropped and uncropped sets. Second, models are trained on both versions of ICDAR 2017, then tested on the corresponding versions of ICDAR 2013. It is worth to note that testing a model on a test set that is completely different from the training set is challenging for the learning models. Dataset and model details are presented in the following subsections.

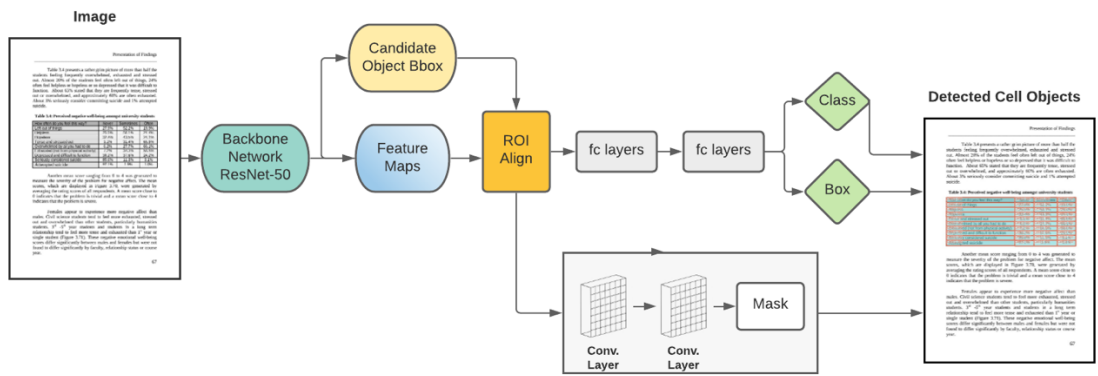

Figure (1) Mask R-CNN architecture for table cell structure detection FC Layers

## 3.2. Networks

Region-based CNN algorithms are widely employed in object detection [20, 21, 22] and have been applied to the problem of detecting table structure. Faster R-CNN has been shown to perform well, and various enhanced versions are proposed for use in combination with advanced backbone architectures. This section delves into these networks in depth.

### 3.2.1. Mask R-CNN with ResNet-50

Mask R-CNN [23] is a two-stage object detector that was built with the addition of a mask branch to Faster R-CNN [24]. The first stage, same as in [24] generates proposal bounding boxes where objects possibly lie via Region Proposal Network (RPN). In the second stage, features are extracted from each region proposal and used to output class prediction and box location. Classification and regression of bounding-box stage were adopted from Fast R-CNN [25]. Mask R-CNN uses RoIAlign instead of RoIPool [25] for feature extraction from RPN outputs. RoIPool causes misalignment between the RoI and extracted features due to the quantization process. RoIAlign solves this problem with a simple change in quantization and aligns the extracted features with the input. We refer readers to [23] for detailed explanations. Fig. 1 illustrates the flow of Mask R-CNN method. ResNet-50 is used as the backbone with Mask R-CNN by following the implementation in [13]. Backbones feed RPN with the extracted features from input images. ResNet-50 has 50 convolutional neural network layers that capture semantic information from low level to high level as it goes into deeper layers. ResNet [26] addresses the degradation of training accuracy problem with increasing depth in networks.

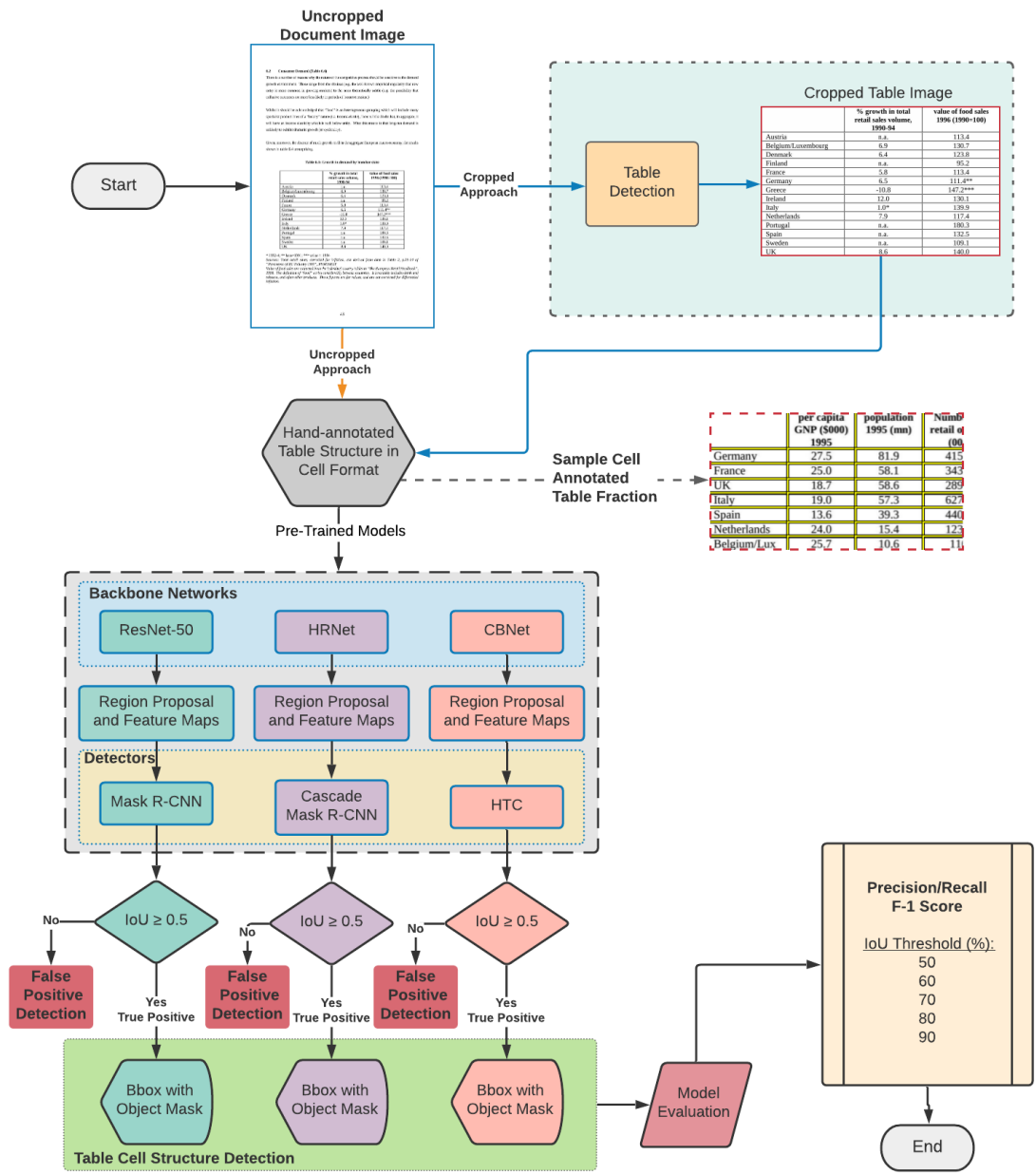

Figure (2) Flow diagram of cropped and uncropped approaches

### 3.2.2. Cascade Mask R-CNN with HRNet

Cascade R-CNN [27] is the extended version of Faster R-CNN with the sequential design of detection head where class and box predictions are made. This cascaded structure refines predictions with a resampling mechanism at each stage. Cascade Mask R-CNN [28] was built with the addition of mask branch in parallel to the detection branch, similar to Mask R-CNN. However, Cascade R-CNN has multiple detection branches. So, authors in [28] investigated different strategies where the segmentation branch is added only to the first stage, last stage or all stages. In this study, we used the one that has mask branch only at the final stage by following the implementation in

[14]. The authors chose HRNetV2 [29] as the backbone network since it provides semantically strong representations in the feature extraction step. Images are downsampled to low resolutions to obtain feature maps and then upsampled to construct high-resolution feature maps. These processes are done by convolutional operations connected in series. HRNetV2 is designed in a multi-stage manner where high-resolution representations are kept, and one lower resolution representation is added at each step in parallel. Besides, information is exchanged between different resolutions by a fusion module that provides semantically rich feature maps with more precise spatial information.

### 3.2.3. Hybrid Task Cascade with CBNet

Hybrid Task Cascade (HTC) [30] was proposed to improve Cascade Mask R-CNN by exploiting the relationship between the detection and segmentation heads. HTC interleaves the box and mask branches instead of performing bounding box predictions and mask predictions in parallel branches. Thus each mask head utilizes the refined bounding box predictions from the next layer. Also, HTC introduces an information flow between the mask predictions at different stages. Mask features at each stage are fed to the next mask branch directly. Finally, a semantic segmentation branch is added in HTC architecture. The segmentation branch provides additional features obtained from Feature Pyramid [31] to the box and mask heads of each stage. This information-sharing design of architecture among the detection, mask and semantic segmentation heads is the key to improvement.We implemented HTC with Double ResNeXt101 backbone as in the work [15]. Composite Backbone Network (CBNet) [32] introduced the idea of composite design of existing backbone networks such as ResNet [33] and ResNeXt [34]. There are two types of backbones, namely the Lead Backbone and the Assistant Backbone in CBNet. The assistant backbone refines the leading backbone outputs at each stage, and refined features are fed into the following stage of the leading backbone. Backbones are called as Dual-Backbone or Triple-Backbone depending on the number of identical backbones assembled.

3.3. Datasets

The publicly available ICDAR 2013 [35], and ICDAR 2017 [36] datasets have been widely used in table detection solutions. In this study, we used these datasets with their cropped and uncropped versions. Images that only contain tables are included in both datasets. The datasets are hand-labelled in cell format for table structure recognition as used in [14, 15]. When table layout is defined in rows and columns format, dealing with the spanning rows or columns is challenging [12, 13]. Cell format is chosen due to its simplicity in defining table structure. The number of images in datasets is summarized in Table 2. Since some of the images have more than one table, cropped sets have more images than uncropped images.

Table (2) Numerical details of used datasets

| Dataset | Cropped | | Uncropped | |
|---|---|---|---|---|
| | Training | Test | Training | Test |
| ICDAR 2013 | - | 156 | - | 128 |
| ICDAR 2017 | 1012 | - | 784 | - |
| ICDAR 2017^ | 806 | 206 | 627 | 157 |

4. Performance Study

This section presents the performance comparison of the copped and uncropped training sets by using the state-of-the-art object detection algorithms that have been proposed for the table

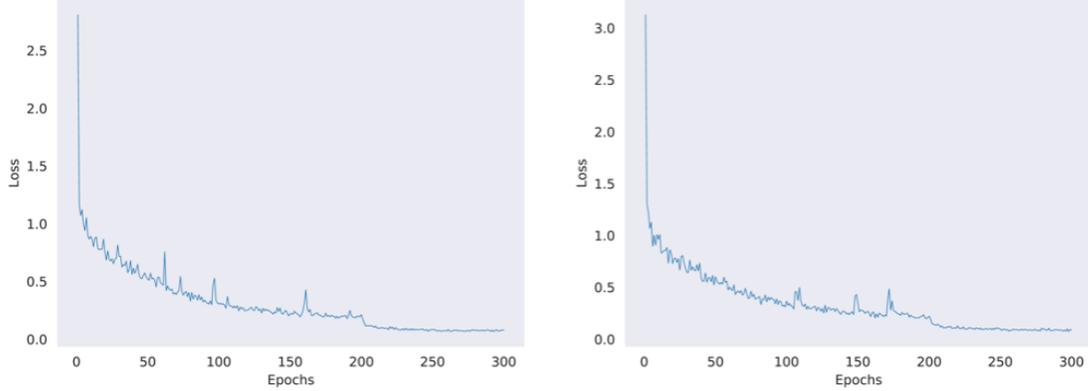

(a) Learning Curve for Mask R-CNN on Cropped Set  (b) Learning Curve for Mask R-CNN on Unropped Set

Figure (3) Learning Curves

structure detection task. There are two different sets of experiments with different datasets. The Mist GPU cluster, which consists of 4 Tesla V 100 GPUs with 32GB VRAM per node, was used in all experiments. All models were implemented by using the mmdetection toolbox [37]. To carry out experiments, Mask R-CNN is chosen as a baseline model. The presented works in [12, 13] has shown that Mask R-CNN performs well on structure detection task. Besides, Mask R-CNN has been widely preferred on instance segmentation due to promising performance and less complex structure. Improving structure detection performance by developing network architecture is not the focus of this study. However, to further investigate the cropped and uncropped sets and justify findings, two more models are used in our experiments. Prasad et al. [14] achieved the highest structure detection results on ICDAR 2019 dataset by using Cascade Mask R-CNN with High-Resolution Network (HRNet) backbone. Jiang et al. [15] created a benchmark study on ICDAR 2013 dataset by using various models with the combination of different backbones. Hybrid Task Cascade (HTC) with double ResNeXt101 outperform all models and shown to be the best candidate for table structure detection. Therefore, Cascade Mask R-CNN HRNet and HTC with dual ResNeXt101 can be chosen for experiments. All models are trained for 300 epochs with both uncropped sets and cropped sets separately. A learning rate of 0.005 was chosen for each model, and training loss and accuracy were watched to assure that the model is fully trained. The learning curves of Mask R-CNN with ResNet-50 for the cropped and uncropped version are presented in Fig. 3a and Fig. 3b respectively.

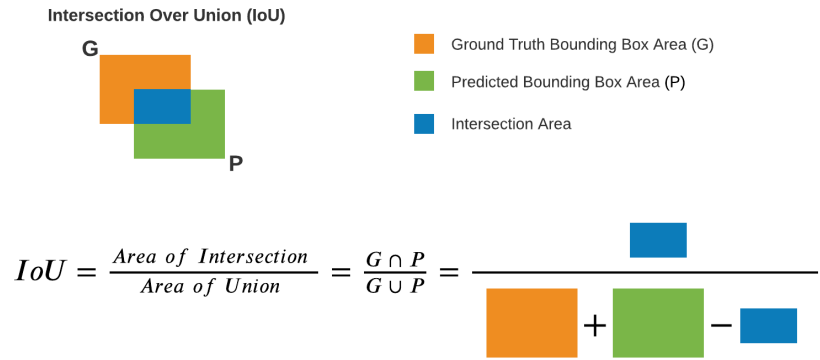

Figure (4) Intersection over Union (IoU) concept

$$IoU = \frac{Area\ of\ Intersection}{Area\ of\ Union} = \frac{G \cap P}{G \cup P}$$

### 4.1. Performance Evaluation

Detections are evaluated based on precision, recall and F1-score. The concept of Intersection over Union (IoU) is used to determine true detections and false detections. IoU is the ratio of overlapping area to combined area of ground truth and predicted bounding box. Evaluation metrics are calculated for 50% IoU threshold and over. If the IoU value is under 50%, detections are considered as false positive. IoU concept is illustrated in Fig. 4.

#### 4.1.1. Experiments with ICDAR 2017 Dataset

In the first set of experiments, the uncropped ICDAR 2017[*] dataset is split into training and test sets in 80% and 20%, respectively. Numerical details are given in Table 2. The cropped training and test sets are created with the corresponding cropped table images. In other words, 806 table images in the cropped ICDAR 2017 dataset are obtained from 627 images in the uncropped training set by cropping tables. The structure detection results of various networks are presented in Table 3, 4, and 5 for the cropped and uncropped versions. All models achieve 6%-9% higher AP, 5%-8% higher AR and 6%-8% higher AF1 consequently on the cropped set. When the results are examined for different IoU thresholds, an interesting relation is observed. Precision and recall values of models are quite close to each other on the cropped and uncropped sets at 50% IoU. Performance difference between these sets can go up to 17% and 15% of precision and recall values, respectively. These are illustrated with precision-IoU plots in Fig. 5 for each model.

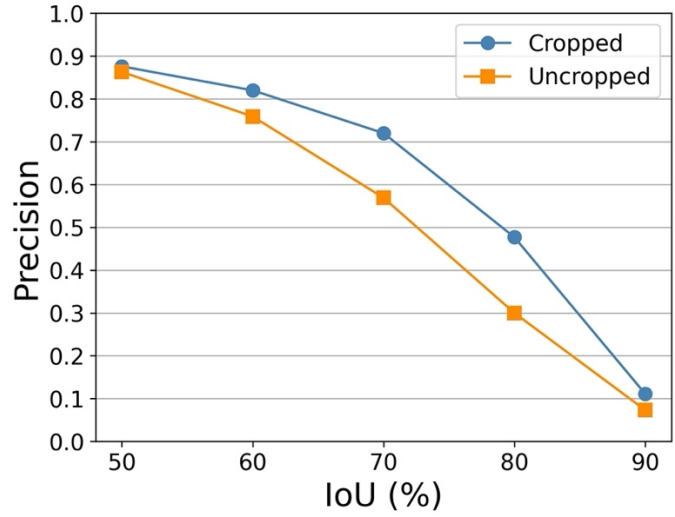

(a) Mask R-CNN

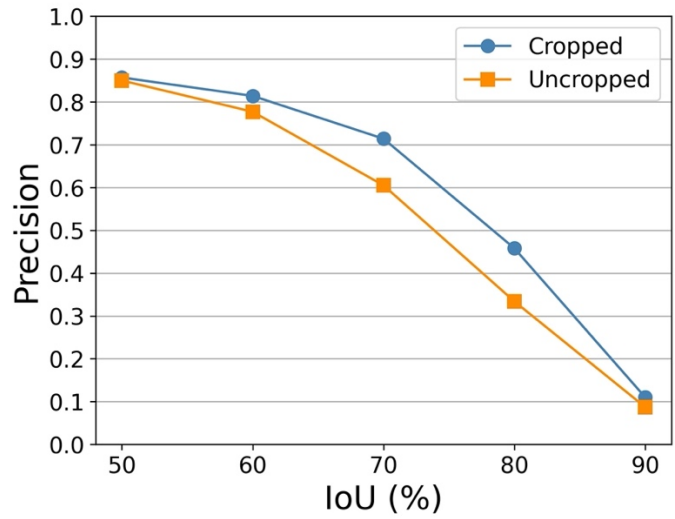

(b) Cascade Mask R-CNN

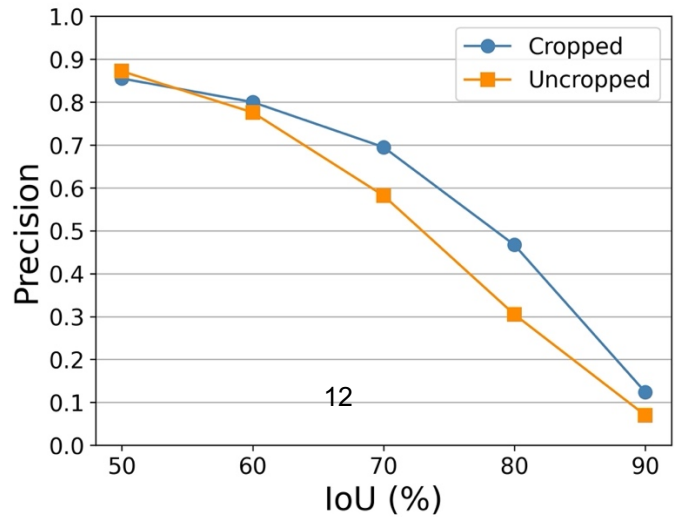

(c) Hybrid Task Cascade

Figure (5) Comparison of precision values of cropped and uncropped ICDAR 2017 datasets with different models

Table (3) Structure Detection Results for The Cropped and Uncropped ICDAR 2017 Datasets with Mask R-CNN

| Metric | Version | Average | IoU 0.50% | 0.60% | 0.70% | 0.80% | 0.90% |
|---|---|---|---|---|---|---|---|
| Precision | Cropped | 0.557 | 0.876 | 0.820 | 0.720 | 0.477 | 0.111 |
| | Uncropped | 0.468 | 0.863 | 0.759 | 0.569 | 0.300 | 0.073 |
| Recall | Cropped | 0.665 | 0.929 | 0.890 | 0.818 | 0.639 | 0.279 |
| | Uncropped | 0.588 | 0.919 | 0.847 | 0.713 | 0.491 | 0.204 |
| F1-Score | Cropped | 0.606 | 0.902 | 0.854 | 0.766 | 0.546 | 0.159 |
| | Uncropped | 0.521 | 0.890 | 0.801 | 0.633 | 0.372 | 0.108 |

Table (4) Structure Detection Results for The Cropped and Uncropped ICDAR 2017 Datasets with Cascade Mask R-CNN

| Metric | Version | Average | IoU 0.50% | 0.60% | 0.70% | 0.80% | 0.90% |
|---|---|---|---|---|---|---|---|
| Precision | Cropped | 0.546 | 0.857 | 0.814 | 0.714 | 0.458 | 0.110 |
| | Uncropped | 0.487 | 0.850 | 0.777 | 0.605 | 0.334 | 0.087 |
| Recall | Cropped | 0.638 | 0.907 | 0.871 | 0.794 | 0.597 | 0.250 |
| | Uncropped | 0.579 | 0.906 | 0.854 | 0.714 | 0.471 | 0.181 |
| F1-Score | Cropped | 0.588 | 0.881 | 0.842 | 0.752 | 0.518 | 0.153 |
| | Uncropped | 0.529 | 0.877 | 0.814 | 0.655 | 0.391 | 0.118 |

Table (5) Structure Detection Results for The Cropped and Uncropped ICDAR 2017 Datasets with Hybrid Task Cascade

| Metric | Version | Average | IoU 0.50% | 0.60% | 0.70% | 0.80% | 0.90% |
|---|---|---|---|---|---|---|---|
| Precision | Cropped | 0.544 | 0.855 | 0.800 | 0.695 | 0.467 | 0.124 |
| | Uncropped | 0.476 | 0.872 | 0.776 | 0.582 | 0.305 | 0.070 |
| Recall | Cropped | 0.646 | 0.918 | 0.874 | 0.789 | 0.601 | 0.280 |
| | Uncropped | 0.597 | 0.932 | 0.869 | 0.727 | 0.493 | 0.200 |
| F1-Score | Cropped | 0.591 | 0.885 | 0.835 | 0.739 | 0.526 | 0.172 |
| | Uncropped | 0.530 | 0.901 | 0.820 | 0.646 | 0.377 | 0.104 |

4.1.2. Evaluation under ICDAR 2013 + ICDAR 2017 Datasets

In the second experiment, the cropped and uncropped version of the ICDAR 2017 dataset is used as the training dataset and trained models tested on the corresponding versions of the ICDAR 2013 dataset. In Table 6, 7, and 8, the structure detection results are given with state-of-the-art table structure detection models. When Mask R-CNN is trained with the cropped dataset, it achieves 0.8% higher AP, 1.4% higher AR and 1.1% higher AF1 than the uncropped set. The uncropped dataset provides higher precision and recall values at 50% and lower IoU values up until 80% IoU. As observed in section 4.1.1, the performance of cropped sets increases with the increasing IoUs.

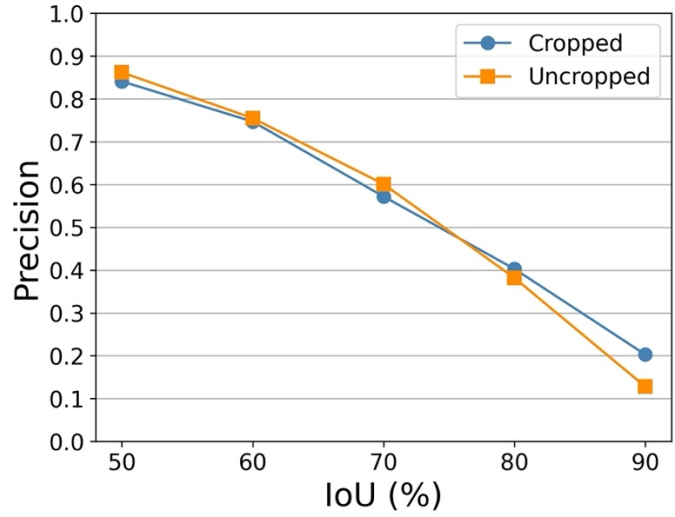

(a) Mask R-CNN

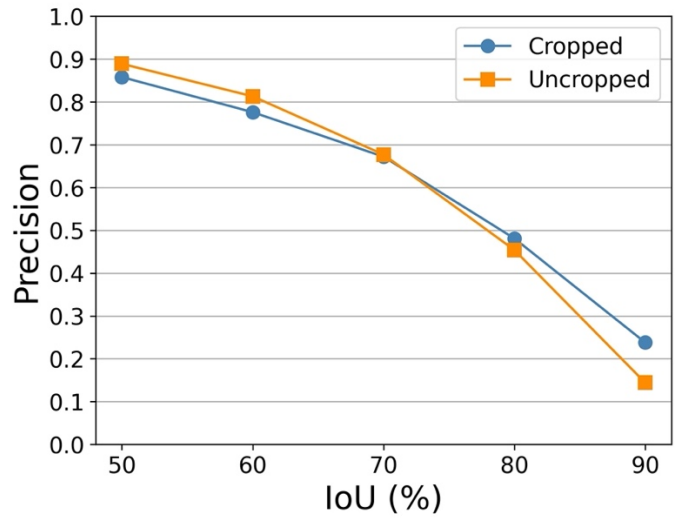

(b) Cascade Mask R-CNN

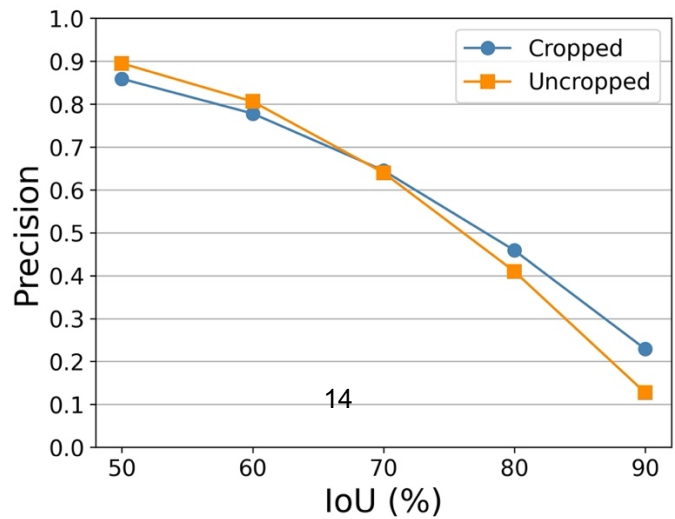

(c) Hybrid Task Cascade

Figure (6) Comparison of precision values of cropped and uncropped ICDAR 2013 datasets with different models

Table (6) Structure Detection Results for The Cropped and Uncropped ICDAR 2013 Datasets with Mask R-CNN

| Metric | Version | Average | IoU 0.50% | 0.60% | 0.70% | 0.80% | 0.90% |
|---|---|---|---|---|---|---|---|
| Precision | Cropped | 0.509 | 0.841 | 0.747 | 0.572 | 0.403 | 0.203 |
| | Uncropped | 0.501 | 0.862 | 0.755 | 0.601 | 0.382 | 0.128 |
| Recall | Cropped | 0.596 | 0.889 | 0.820 | 0.673 | 0.505 | 0.310 |
| | Uncropped | 0.582 | 0.894 | 0.820 | 0.696 | 0.499 | 0.232 |
| F1-Score | Cropped | 0.549 | 0.864 | 0.782 | 0.618 | 0.448 | 0.245 |
| | Uncropped | 0.538 | 0.878 | 0.786 | 0.645 | 0.433 | 0.165 |

Table (7) Structure Detection Results for The Cropped and Uncropped ICDAR 2013 Datasets with Cascade Mask R-CNN

| Metric | Version | Average | IoU 0.50% | 0.60% | 0.70% | 0.80% | 0.90% |
|---|---|---|---|---|---|---|---|
| Precision | Cropped | 0.559 | 0.858 | 0.776 | 0.672 | 0.481 | 0.238 |
| | Uncropped | 0.549 | 0.889 | 0.813 | 0.677 | 0.454 | 0.144 |
| Recall | Cropped | 0.611 | 0.881 | 0.819 | 0.724 | 0.543 | 0.309 |
| | Uncropped | 0.614 | 0.911 | 0.854 | 0.749 | 0.556 | 0.235 |
| F1-Score | Cropped | 0.584 | 0.869 | 0.797 | 0.697 | 0.510 | 0.269 |
| | Uncropped | 0.580 | 0.900 | 0.833 | 0.711 | 0.500 | 0.179 |

Table (8) Structure Detection Results for The Cropped and Uncropped ICDAR 2013 Datasets with Hybrid Task Cascade

| Metric | Version | Average | IoU 0.50% | 0.60% | 0.70% | 0.80% | 0.90% |
|---|---|---|---|---|---|---|---|
| Precision | Cropped | 0.548 | 0.859 | 0.778 | 0.645 | 0.459 | 0.229 |
| | Uncropped | 0.529 | 0.895 | 0.806 | 0.640 | 0.410 | 0.127 |
| Recall | Cropped | 0.631 | 0.910 | 0.851 | 0.736 | 0.556 | 0.324 |
| | Uncropped | 0.628 | 0.939 | 0.874 | 0.747 | 0.554 | 0.263 |
| F1-Score | Cropped | 0.587 | 0.884 | 0.813 | 0.688 | 0.503 | 0.268 |
| | Uncropped | 0.574 | 0.916 | 0.839 | 0.689 | 0.471 | 0.171 |

Comparison between cropped and uncropped datasets is further investigated with Cascade Mask R-CNN and HTC. Cascade Mask R-CNN also provides one percent better AP and 0.4% AF1 score with the cropped set, while the average AR value is 0.3% lower than the uncropped set. Similar to Mask R-CNN, Cascade Mask R-CNN performs well at higher IoUs on the cropped set. The AP value of the cropped set is 1.9% improved compared to the uncropped version with the HTC model. The same trend in evaluation metrics between the cropped and uncropped versions for lower and higher IoUs is observed with the HTC model as well. This trend can be seen in Fig. 6 for each model.

Overall, the structure detection performance of all models is improved on the cropped table images for AP, AR and AF1 metrics. Models either provided better cell structure detection based on the evaluation metrics under lower IoU values on the uncropped table images or performed similarly with the cropped sets. However, with the increasing IoU thresholds, models achieve significantly higher AP, AR and AF1 values on cropped sets. In other words, cell structures are detected more accurately on cropped table images. Cell structure detection samples are presented in Appendix A for uncropped table images and corresponding cropped table images.

5. Conclusion

Despite lacking analysis on the interplay between them, table detection and table structure recognition are have been considered two consecutive tasks. In this paper, we have investigated the table cell structure detection performance on the cropped and uncropped versions of the ICDAR 2013 and ICDAR 2017 datasets. Comparison of these two versions illustrates the impact of having a table detection model. By proving that the cropped version provides remarkably better performance, it has become a guide for researchers in future table structure detection studies. Experiments are initially done by using Mask R-CNN. Cascade Mask R-CNN and Hybrid Task Cascade are used for further analysis. Models achieved higher AP and AR on the cropped datasets. Besides, the following impact of the IoU thresholds on model performance has been been discovered: Models can provide better detections at lower IoU values of 50%-70% on either set, while they perform noticeably better under higher IoU values on the cropped set. The performance gap can be as wide as 15% and 17% in terms of the precision and recall values. It can be concluded that none-table objects in document images such as textual areas, figures and plots degrade the model performance for higher IoU thresholds. Hence a robust table detection model improves the performance of a table structure detection model. Finally, false-positive detections due to lines existing in figures or alignments in textual areas can be eliminated by the cropping process.

Appendix A. Inference Results

To complement numerical results in the study, this Appendix section provides sample inference results on ICDAR2013 dataset. Figure A.7a and A.7c show that lines in chart figures within the documents cause false positive detections. On croppped table images, these false detections can be eliminated as seen in Figure A.7b and A.7d.

Furthermore, vertical or horizontal alignments of the textual region might be perceived as tabular information. By working on cropped sets, these deceiving factors can be reduced to minimum. A sample image that shows false positive cell object detections in a textual area can be seen in Figure A.7e and the corresponding cropped table structure detections in A.7f.

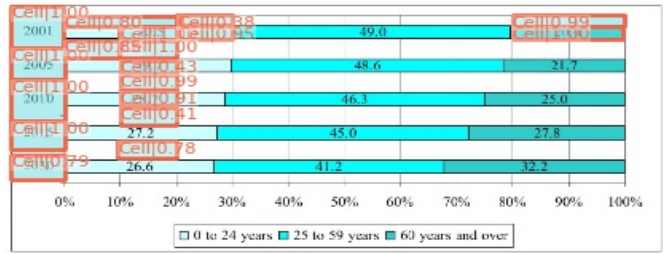

Figure 1: Population by age group in 2001, 2005 and forecasts for 2010, 2015 and 2030 (in %)

Source: Statistics Finland, 2006.

## 1.3. The economy and the labour market

The economy and welfare have grown steadily in Finland since independence until the 1990s, except during the depression in the 1930s and the Second World War. In the 1950s, trade with the Soviet Union had a significant impact on the development of export industries. First the war indemnities to Soviet Union and then the bilateral trade relations with it meant a rapid increase in industrial activity in Finland. In the 1980s growth was stable but, at the beginning of the 1990s, the Finnish national economy was hit by the worst depression since the war. The growth of GDP in recent years has been faster than in the EU in general (see Table 1).

Table 1: Real GDP growth rate in Finland, EU-15 and EU-25 for 1996, 2000, 2005 and 2006 (percentage change on previous year)

|  | Finland | EU-15 | EU-25 |
|---|---|---|---|
| 1996 | 3.7 | 1.6 | 1.7 |
| 2000 | 5.0 | 3.9 | 3.9 |
| 2005 | 1.5 | 1.5 | 1.6 |
| 2006 (*) | 3.5 | 2.0 | 2.1 |

GDP: Gross domestic product.
(*) Forecast.
Source: Eurostat. European system of accounts (ESA 1995), 2005.

Finland has the industrial structure of a modern knowledge-based society. The proportion of agriculture and manufacturing has declined and, in the last two decades, electronics has become the success story of Finnish exports. Its growth in the 1990s is mainly based on mobile phones and other telecommunication equipment. Three major export sectors today are

(a)

|  | Finland | EU-15 | EU-25 |
|---|---|---|---|
| 1996 | 3.7 | 1.6 | 1.7 |
| 2000 | 5.0 | 3.9 | 3.9 |
| 2005 | 1.5 | 1.5 | 1.6 |
| 2006 (*) | 3.5 | 2.0 | 2.1 |

(b)

155. Specific events and factors were of particular importance in the decline of ABCPs. Firstly, some conduits had large ABS holdings that experienced huge declines. When investors stopped rolling over ABCPs, these conduits had to rely on guarantees provided by banks which were too large for the banks providing them. While these banks received support to meet their obligations, investor confidence was nonetheless damaged. Secondly, structures in other ABCP markets around the world unsettled investors, including different guarantee agreements and single-seller extendible mortgage conduits. Thirdly, general concerns about the banking sector have caused investors to buy less bank related product.

**Table 3** - European ABCP issuance

|      | Q1    | Q2   | Q3    | Q4    | Total |
|------|-------|------|-------|-------|-------|
| 2004 | 34.7  | 36.2 | 44.5  | 51.3  | 166.7 |
| 2005 | 58.1  | 63.4 | 61.6  | 55.2  | 238.4 |
| 2006 | 74.7  | 84.1 | 96.5  | 111.8 | 367.1 |
| 2007 | 148.8 | 142.3| 156.7 | 186.1 | 633.9 |
| 2008 | 120.9 | 106  |       |       | 226.8 |

Source: Moody's, Dealogic, ESF

Chart 5

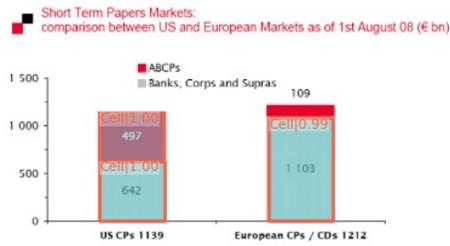

Source: Société Générale Corporate & Investment Banking (market overview, 19 September, 2008)

**Credit Derivatives Markets**

156. The credit derivatives markets comprise a number of instruments. Credit default swaps represent, by far, the single most significant credit derivative instrument in terms of volume. Other credit derivative instruments are not covered in this consultation paper[43].

---
[43] Examples of credit derivatives not included in the scope of this consultation paper are total return swaps and credit linked notes.

(c)

|      | Q1    | Q2   | Q3    | Q4    | Total |
|------|-------|------|-------|-------|-------|
| 2004 | 34.7  | 36.2 | 44.5  | 51.3  | 166.7 |
| 2005 | 58.1  | 63.4 | 61.6  | 55.2  | 238.4 |
| 2006 | 74.7  | 84.1 | 96.5  | 111.8 | 367.1 |
| 2007 | 148.8 | 142.3| 156.7 | 186.1 | 633.9 |
| 2008 | 120.9 | 106  |       |       | 226.8 |

(d)

(e)

(f)

Figure (A.7)    Sample structure detection results on cropped and uncropped ICDAR 2013 datasets.